\documentclass[preprint,5p,times,twocolumn]{elsarticle}

\usepackage{amssymb}

\usepackage[nodots]{numcompress}

\usepackage{amsfonts} 

\usepackage[pagebackref=true,breaklinks=true,letterpaper=true,colorlinks,bookmarks=false]{hyperref}

\usepackage{color}
\usepackage[utf8]{inputenc}
\usepackage[T1]{fontenc}
\usepackage{times}
\usepackage{epsfig}
\usepackage{amsmath}
\usepackage{amssymb}

\usepackage{caption}
\usepackage{subcaption}
\usepackage{algpseudocode,algorithm,algorithmicx}

\usepackage{mathptmx}      

\journal{Springer}

\renewcommand{\eqref}[1]{Equation~(\ref{#1})}

\makeatletter
\newcommand{\Spvek}[2][r]{%
	\gdef\@VORNE{1}
	\left(\hskip-\arraycolsep%
	\begin{array}{#1}\vekSp@lten{#2}\end{array}%
	\hskip-\arraycolsep\right)}

\def\vekSp@lten#1{\xvekSp@lten#1;vekL@stLine;}
\def\vekL@stLine{vekL@stLine}
\def\xvekSp@lten#1;{\def\temp{#1}%
	\ifx\temp\vekL@stLine
	\else
	\ifnum\@VORNE=1\gdef\@VORNE{0}
	\else\@arraycr\fi%
	#1%
	\expandafter\xvekSp@lten
	\fi}
\makeatother
\begin{document}

	\begin{frontmatter}



		\title{Fast camera focus estimation for gaze-based focus control}

		

		\author[adrtue]{Wolfgang Fuhl}
		\author[adrtue]{Thiago Santini}
		\author[adrtue]{Enkelejda Kasneci}

		\address[adrtue]{Eberhard Karls University T\"ubingen, Perception Engineering, Germany, 72076 T\"ubingen, Sand 14, Tel.: +49 70712970492, wolfgang.fuhl@uni-tuebingen.de, thiago.santini@uni-tuebingen.de, Enkelejda.Kasneci@uni-tuebingen.de}

		\begin{abstract}
Many cameras implement auto-focus functionality; however, they typically require
the user to manually identify the location to be focused on. While such an approach works for temporally-sparse autofocusing functionality (e.g., photo
shooting), it presents extreme usability problems when the focus must be quickly
switched between multiple areas (and depths) of interest -- e.g., in a gaze-based autofocus approach.
This work introduces a novel, real-time auto-focus approach based on
eye-tracking, which enables the user to shift the camera focus plane swiftly
based solely on the gaze information.
Moreover, the proposed approach builds a graph representation of the image to estimate
depth plane surfaces and runs in real time (requiring $\approx20ms$ on a single
i5 core), thus allowing for the depth map estimation to be performed
dynamically.
We evaluated our algorithm for gaze-based depth estimation against state-of-the-art approaches
based on eight new data sets with flat, skewed, and round surfaces, as well as
publicly available datasets.
\end{abstract}


	\end{frontmatter}



\section{Introduction}
Human vision is a foveated vision, i.e., sharpest vision is possible only in the central $2^\circ$ of the field of view. Therefore, in order to perceive our environment, we perform eye movements, which allow us to focus on and switch between regions of interest sequentially~\cite{Tafaj2012}. Modern digital microscopes and cameras share a similar problem, in the sense that the auto focus is only applied to the center of the field of view of the camera.

For various applications -- such as microsurgery, microscopical material inspection, human-robot collaborative settings, security cameras -- it would be a significant usability improvement to allow users to adjust the focus in the image to their point of interest without reorienting the camera or requiring manual focus adjustments. This would not only generate a benefit for the user of the optical system, but also to non-users -- for instance, patients would benefit from a faster surgery and a less strained surgeon. For a security camera, the security staff could watch over a complete hall with eye movement speed, quickly scanning the environment for suspicious activity. Applied to different monitors, the efficiency gain would be even greater.

In this work, we used a commercial eye tracker to capture the subject's gaze. This gaze is then mapped to a screen where the camera images (two images are overlaid in red and cyan for a 3D impression) are presented. The gaze position on the image is mapped to the estimated depth map, from which the focal length of the camera is automatically adjusted. This enables the user to quickly explore the scene without manually adjusting the camera's focal length. The depth map creation for the complete scene takes $\approx$20ms (single core from i5), whereas the preprocessing for each image takes $\approx$15ms. For depth map creation, we record 20 images with different focal lengths, although this process can be completed with fewer images by trading off accuracy. This leads to a system update time of 332ms based on our cameras frame rate (60Hz or 16.6ms per frame), but buffering delays increase this time to $\approx$450ms. It is worth noticing that this process can be sped-up through a faster camera, multiple cameras, and GPU/multicore processing but is limited by the reaction time of the optotune lens (2.5ms~\cite{datasheetel1030}), which is used to change the focus.

In the following sections, we use depth as an index in relation to the acquired image set. The real world depth can be calculated using the resulting focal length (acquired from the index) and the camera parameters.

\section{Related work}
To compute a 3D representation of an area given a set of images produced with different camera focal lengths, two steps have to be applied. The first step is measuring or calculating how focused each pixel in this set is. Afterwards, each pixel is assigned an initial depth value (usually the maximum response) on which the 3D reconstruction is performed.\\
In the sequence, we describe the shape-from-focus measuring operators briefly, grouping then similarly to Pertuz et al.~\cite{pertuz2013analysis} -- which we refer the reader to for a wider overview.
\begin{itemize}
	\item \textbf{Gradient}-based measure operators are first or higher order derivatives of the Gaussian and are commonly applied for edge detection. The idea here is that unfocused or blurred edges have a lower response than sharp or focused edges. The best performing representatives according to \cite{pertuz2013analysis} are first-order derivatives \cite{russell2007evaluation,geusebroek2000robust}, second central moment on first order derivatives \cite{pech2000diatom}, and the second central moment on a Laplacian (or second order derivatives) \cite{pech2000diatom}.
	\item \textbf{Statistics}-based measurements are based on calculated moments of random variables. These random variables are usually small windows shifted over the image. The idea behind statistics for focus measurements is that the moments (especially the second central moment) reach their maximum at focused parts of the image. According to \cite{pertuz2013analysis}, the best representatives are \cite{yap2004image} using Chebyshev moments, second central moment of the principal components obtained from the covariance matrix \cite{wee2007measure}, second central moment of second central moments in a window \cite{pech2000diatom}, and second central moment from the difference between the image and a blurred counterpart \cite{groen1985comparison,santos1997evaluation,sun2004autofocusing,huang2007evaluation}.
	\item \textbf{Frequency}-based measures transform the image to the frequency domain, which is usually used in image compression. These transformations are\\ Fourier, wavelet or curvelet transforms. Afterwards, the coefficients of the base functions are summed \cite{yang2003wavelet,xie2006wavelet,huang2005robust} or the statistical measures are applied on the coefficients \cite{pertuz2013analysis}. The idea behind frequency-based focus measure is that the need for many base functions (or non zero coefficients) to describe an image is a measure of complexity or structure in the image. This amount of structure or complexity is the measure of how focused the image is.
	\item \textbf{Texture}-based measures use recurrence of intensity values \cite{hilsenstein2005robust,santos1997evaluation,sun2004autofocusing}, computed locally binary patterns \cite{lorenzo2008exploring}, or the distance of orthogonally computed gradients in a window \cite{huang2007evaluation}. The idea here is equivalent to the frequency-based approaches, meaning that the amount of texture present (complexity of the image) is the measure of how focused the image is.
\end{itemize}
Regarding 3D reconstructions, common methods are:
\begin{itemize}
	\item \textbf{Gaussian and polynomial fit}: These techniques fit a Gaussian~\cite{nayar1994shape} or polynomial~\cite{subbarao1995accurate} to the set of focus measures. To accomplish this, samples are collected outgoing from the maximum response of a pixel in the set of measurements (for each image, one measurement) in both directions. The maximum of the resulting Gaussian or polynomial is then used as depth estimate.
	\item \textbf{Surface fitting}: Here the samples for the fitting procedure are volumes around a pixel of focus measures. The surface is fitted to those samples, and the value aligned (in direction of the set of measurements) to the pixel is used as a new value. The improvement to the Gaussian or polynomial fit is that the neighborhood of a pixel influences its depth estimation too. This approach together with a neuronal network for final optimization has been proposed by \cite{asif2001shape}.
	\item \textbf{Dynamic programming}: In this approach, the volume is divided into subvolumes. For each sub-volume, an optimal focus measure based on the result of a least squares optimization technique is computed. These results are combined and used as depth estimation~\cite{ahmad2005heuristic,ahmad2006shape,suwajanakorn2015depth,moeller2015variational}.    \item \textbf{Surface fitting}: Here the samples for the fitting procedure are volumes around a pixel of focus measures. The surface is fitted to those samples, and the value aligned (in direction of the set of measurements) to the pixel is used as a new value. The improvement to the Gaussian or polynomial fit is that the neighborhood of a pixel influences its depth estimation too. This approach together with a neuronal network for final optimization has been proposed by \cite{asif2001shape}.
\end{itemize}

\section{Setup description}
\begin{figure}
	\centering
	\includegraphics[width=0.45\textwidth]{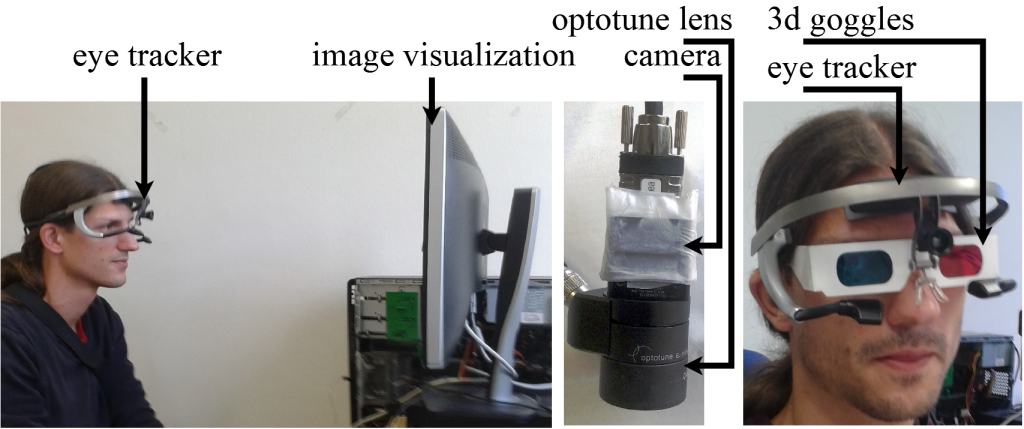}
	\caption{The system for image recording consisting of a digital camera (XIMEA mq013mg\-e2) and an optotune lens (el\-10\-30). On the left side, the subject with eye tracker looking at the image visualization is shown. The same subject with 3D goggles is shown on the right.}
	\label{fig:system}
\end{figure}
As shown in Figure~\ref{fig:system}, the setup consists of a binocular Dikablis Professional eye tracker~\cite{ergoneers2016}, a conventional computer visualizing the image from the camera, and an optotune lens. The optotune lens has a focal tuning range of 50mm to 120mm~\cite{datasheetel1030}, which can be adjusted online over the lens driver (serial communication). The reaction time of the lens is 2.5ms~\cite{datasheetel1030}. We used the XIMEA mq013mg\-e2 digital camera shown in Figure~\ref{fig:system} with a frame rate of 60 Hz and resolution 1280x1024 (we used the binned image 640x512).

For estimating the subjects gaze we used the software EyeRec~\cite{TWTE022016} and the pupil center estimation algorithm ElSe~\cite{WTTE032016}. The calibration was performed with a nine-point grid, fitting a second order polynomial to the pupil center in the least squares sense.

Additionally, it has to be noted that our setup includes a set of fixed lenses between the optotune lens and the object, which could not be disclosed at the time of writing (due to a Non-Disclosure Agreement).

\section{Application}
\begin{figure}
	\centering
	\includegraphics[width=0.45\textwidth]{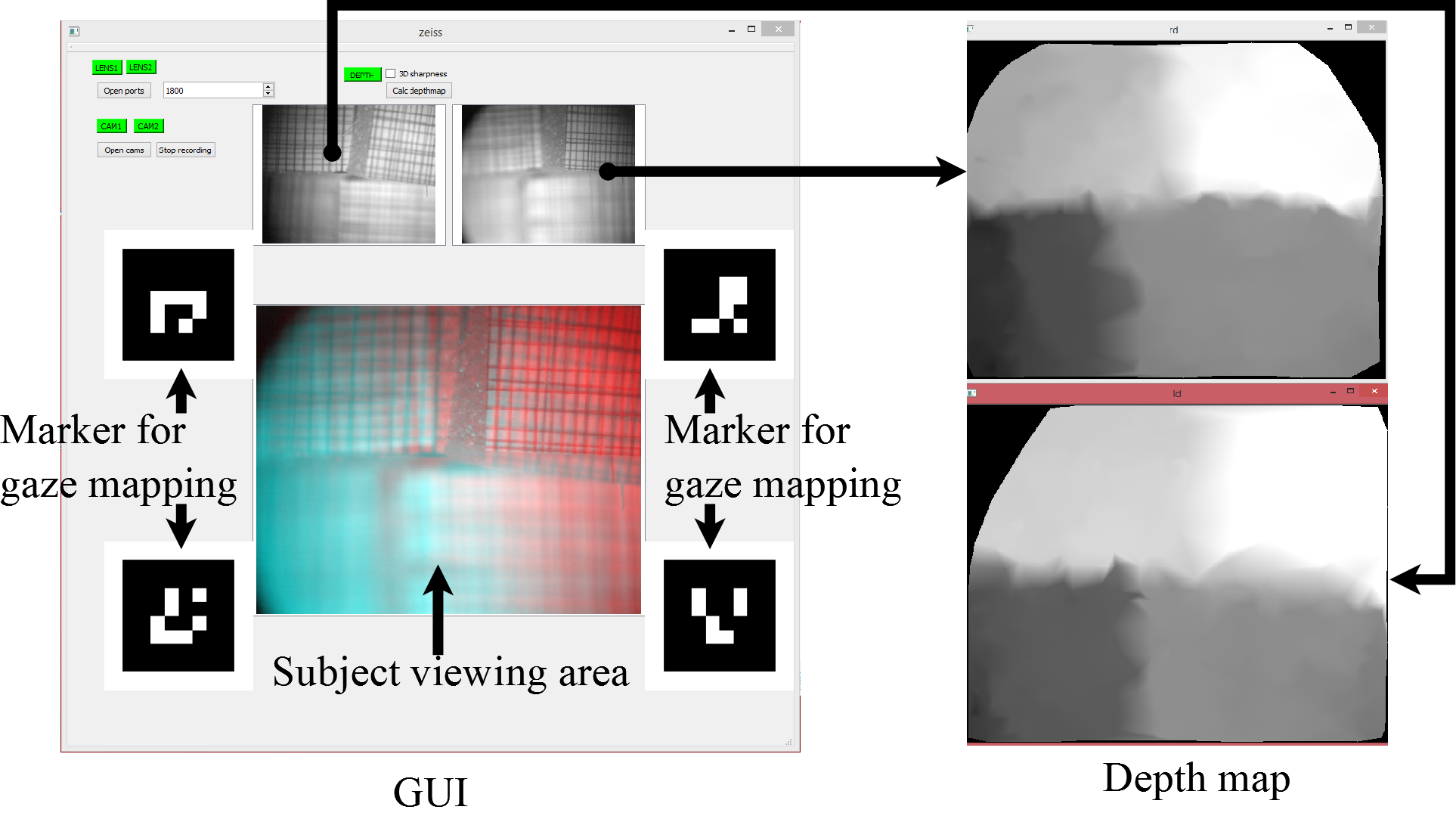}
	\caption{The GUI of the system. In the top row, the images from the two cameras with optotune lenses are shown. The depth map for those is on the right. The correspondence is marked by an arrow. Markers on the left side are used to map the gaze coordinates from the head-mounted eye tracker to the subjects view area. For a 3D representation to the user, we overlay the images from both cameras in red and cyan, which can be seen in the subjects viewing area. }
	\label{fig:anwendung}
\end{figure}
The Graphical User Interface (GUI) of the system can be seen in Figure~\ref{fig:anwendung}; in this GUI, the subjects gaze is mapped to the viewing area through marker detection and transformation between the eye tracker coordinates to screen coordinates. The top two images are from two cameras with optotune lenses. Their depth maps can be seen on the right side and the correspondence is indicated by an arrow. Based on a slight shift of both cameras it is possible to use the red-cyan technique, to achieve a 3D impression for the user too (Figure~\ref{fig:system} on the right side). The focal length of both cameras is automatically set to the depth at the users gaze position.

\section{Method}
\begin{figure}
	\centering
	\includegraphics[width=0.45\textwidth]{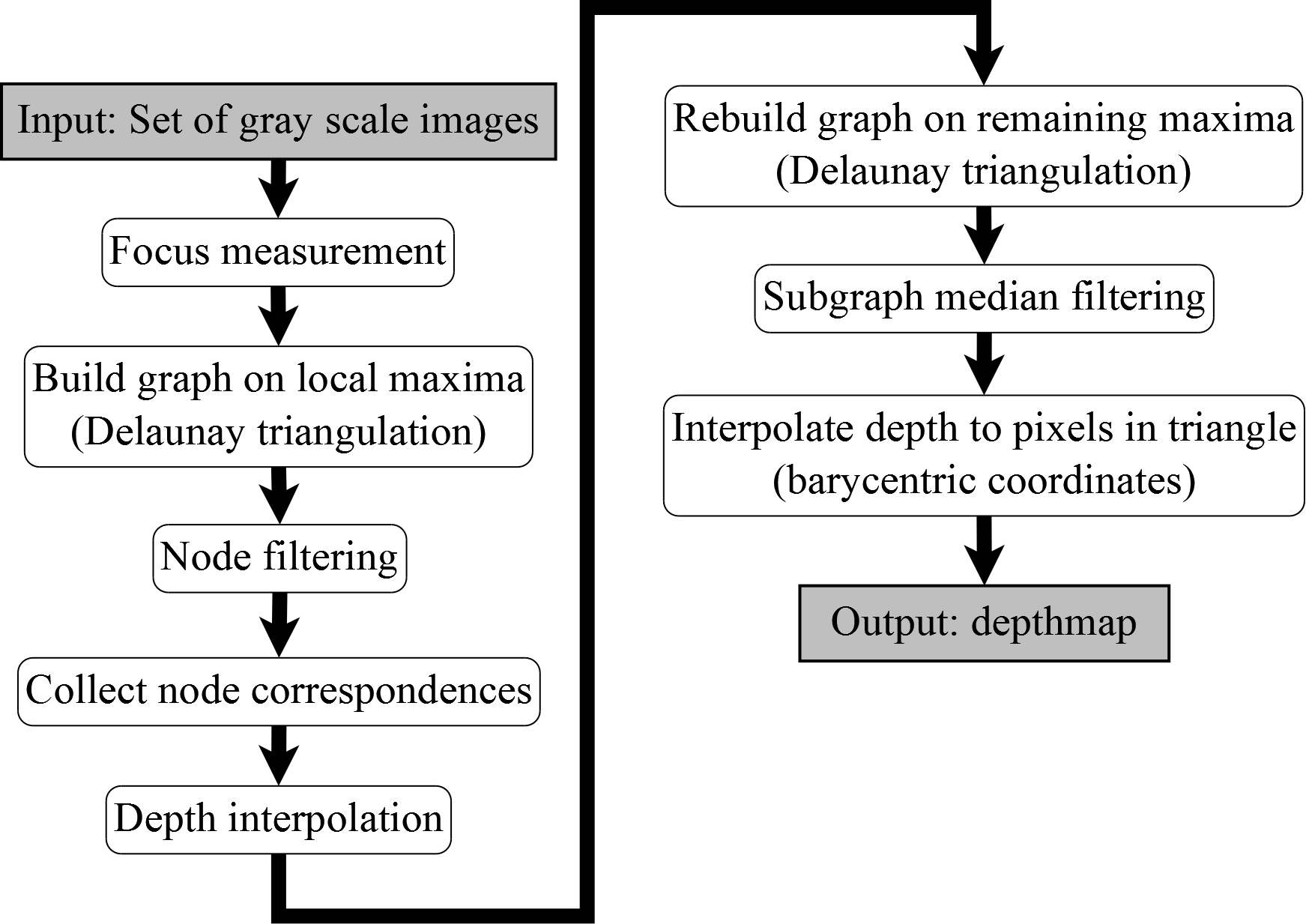}
	\caption{The algorithmic workflow. The gray boxes are in an output of the algorithm. White boxes with rounded corners are algorithmic steps.}
	\label{fig:ablauf}
\end{figure}
All steps of the algorithm are shown in Figure~\ref{fig:ablauf}. The input to the algorithm is a set of grayscale images recorded with different focal length. The images have to be in the correct order otherwise the depth estimation will assign wrong depth values to focused parts of the volume.
\begin{figure}
	\centering
	\begin{subfigure}[b]{0.22\textwidth}
		\includegraphics[width=\textwidth]{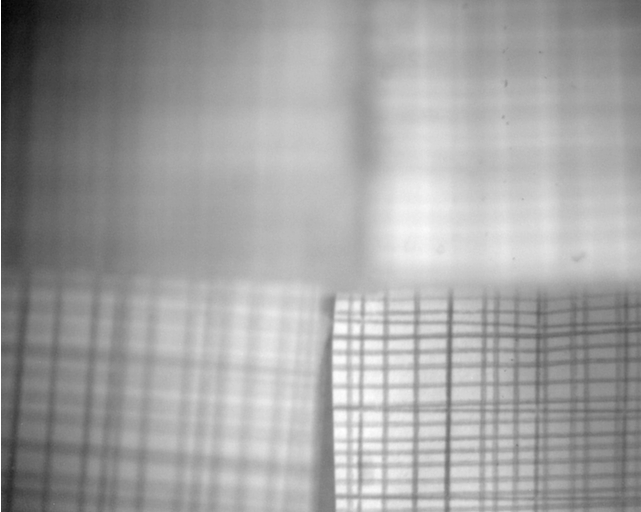}
		\caption{Input}
		\label{fig:focus_input}
	\end{subfigure}
	\begin{subfigure}[b]{0.22\textwidth}
		\includegraphics[width=\textwidth]{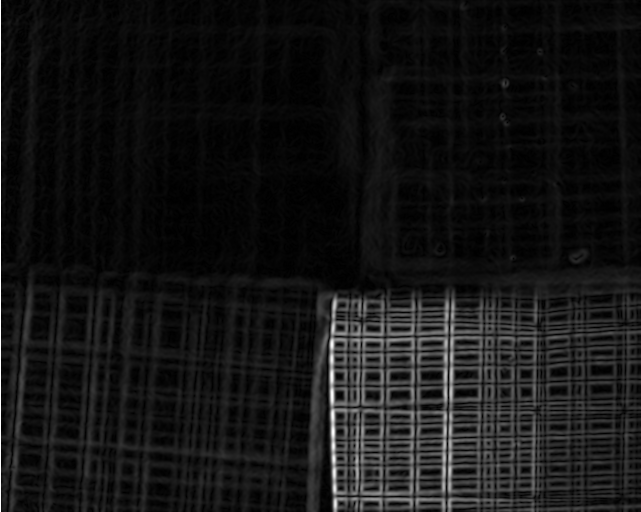}
		\caption{Magnitude}
		\label{fig:focus_mag}
	\end{subfigure}
	
	\begin{subfigure}[b]{0.22\textwidth}
		\includegraphics[width=\textwidth]{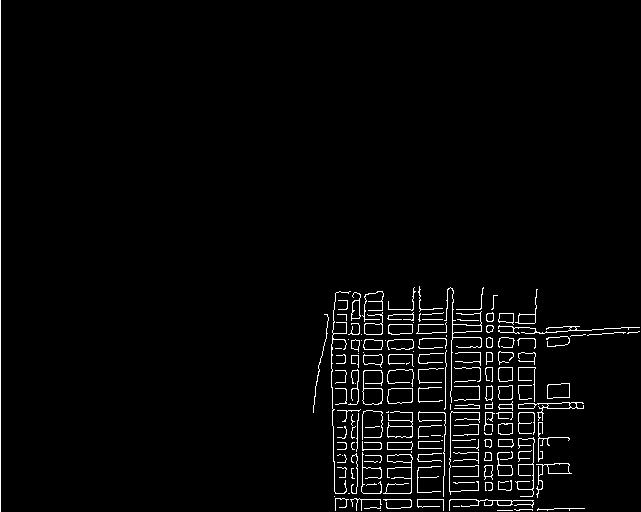}
		\caption{Edges}
		\label{fig:focus_edges}
	\end{subfigure}
	\begin{subfigure}[b]{0.22\textwidth}
		\includegraphics[width=\textwidth]{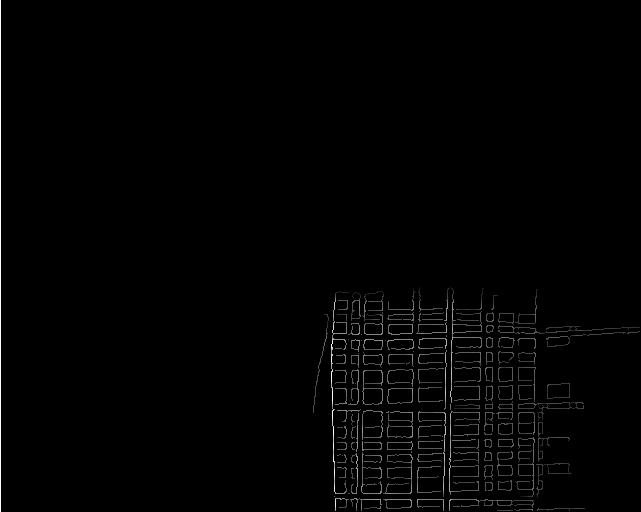}
		\caption{Filtered magnitude}
		\label{fig:focus_mag_final}
	\end{subfigure}
	\caption{Canny edge based in focus estimation for one input image~\ref{fig:focus_input}. In \ref{fig:focus_mag} and \ref*{fig:focus_edges} the output of the canny edge filter is shown and the filtered magnitude image in \ref{fig:focus_mag_final}.}
	\label{fig:Focus_measurement}
\end{figure}
The main idea behind the algorithm is to estimate the depth map only based on parts of the image in which the focus is measurable, and interpolate it to the surrounding pixels if possible.
Regions, where the focus is measurable, are clear edges or texture in an image. Plain regions, for example, usually induce erroneous depth estimations, which have to be filtered out afterward, typically using a median filter in classical shape-from-focus methods. Therefore, we use the Canny edge detector~\cite{canny1986computational} as focus measure. The applied filter is the first derivative of a Gaussian. The resulting edges are used to filter the magnitude response of the filter, allowing only values with assigned edges to pass. For each filtered pixel magnitude, a maximum map through the set of responses is collected. In this map, most of the pixels have no value assigned. Additionally, it has to be noticed that the same edge can be present in this map multiple times because the changing focal length influences the field of view of the camera. This leads to tracing edges in the maximum map.

After computing and filtering the focus measures of the image set, they have to be separated into parts. Therefore, we need candidates representing a strong edge part and their corresponding edge trace to interpolate the depth estimation for the candidate pixel. The candidate selection is performed by only selecting local maxima using an eight-connected neighborhood in the maximum map. These local maxima are used to build a graph representing the affiliation between candidates. This graph is build using the Delaunay triangulation, connecting candidates without intersections.

The separation of this graph into a maximum response and edge trace responses is performed by separating nodes that are maximal in their neighborhood from those that are not. For interpolation of the depth value of maximal nodes, nonmaximal nodes are assigned based on their interconnection to the maxima and to an additional nonmaximal node. Additionally the set of responses is searched for values at the same location since the influence of the field of view does not affect all values in the image, and, as a result, centered edges stay at the same location. The interpolation is performed fitting a Gaussian (as in \cite{nayar1994shape}) to all possible triple assignments and using the median of all results.

The graph spanned by the maxima nodes and the corresponding interpolated depth values are now the representation of the depth map. For further error correction, an interdependent median filter is applied to each node and its direct neighbors in a non-iterative way to ensure convergence. The last part of the algorithm is the conversion of this graph into a proper depth map. It has to be noticed that this graph is a set of triangles spanned between maximal nodes. Therefore, each pixel in the resulting depth map can be interpolated using its barycentric coordinates between the three assigned node values of the triangle it is assigned to. Pixels not belonging to a triangle have no assigned depth value. All steps are described in the following subsections in more detail.

\subsection{Focus Measurement}
\begin{figure}
	\centering
	\begin{subfigure}[b]{0.22\textwidth}
		\includegraphics[width=\textwidth]{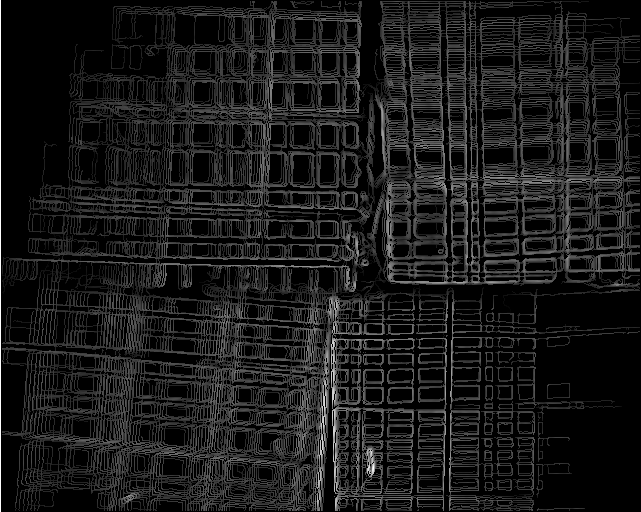}
		\caption{Magnitude of maximum values}
		\label{fig:max_magni}
	\end{subfigure}
	\begin{subfigure}[b]{0.22\textwidth}
		\includegraphics[width=\textwidth]{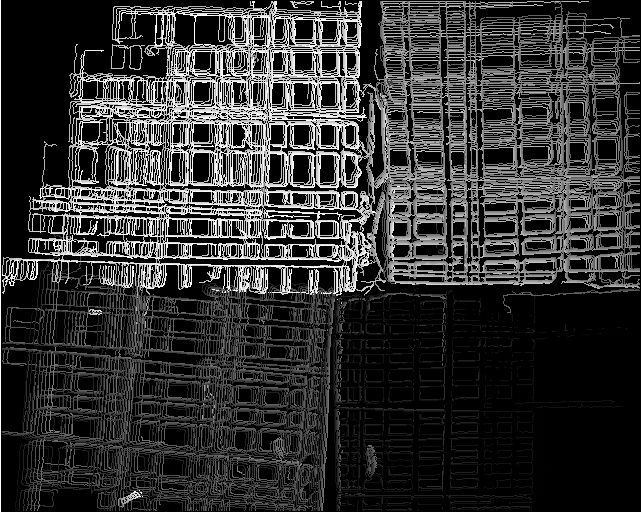}
		\caption{Depth of maximum values ~~~~~~~~~~~}
		\label{fig:max_depth}
	\end{subfigure}
	
	\begin{subfigure}[b]{0.22\textwidth}
		\includegraphics[width=\textwidth]{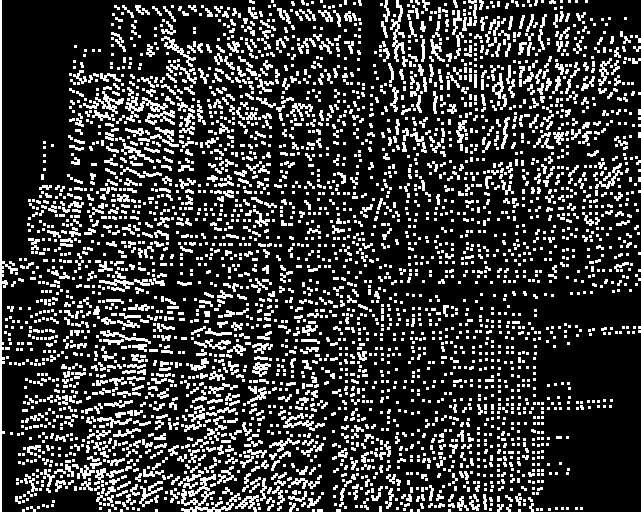}
		\caption{Local maxima}
		\label{fig:max_locmax}
	\end{subfigure}
	\begin{subfigure}[b]{0.22\textwidth}
		\includegraphics[width=\textwidth]{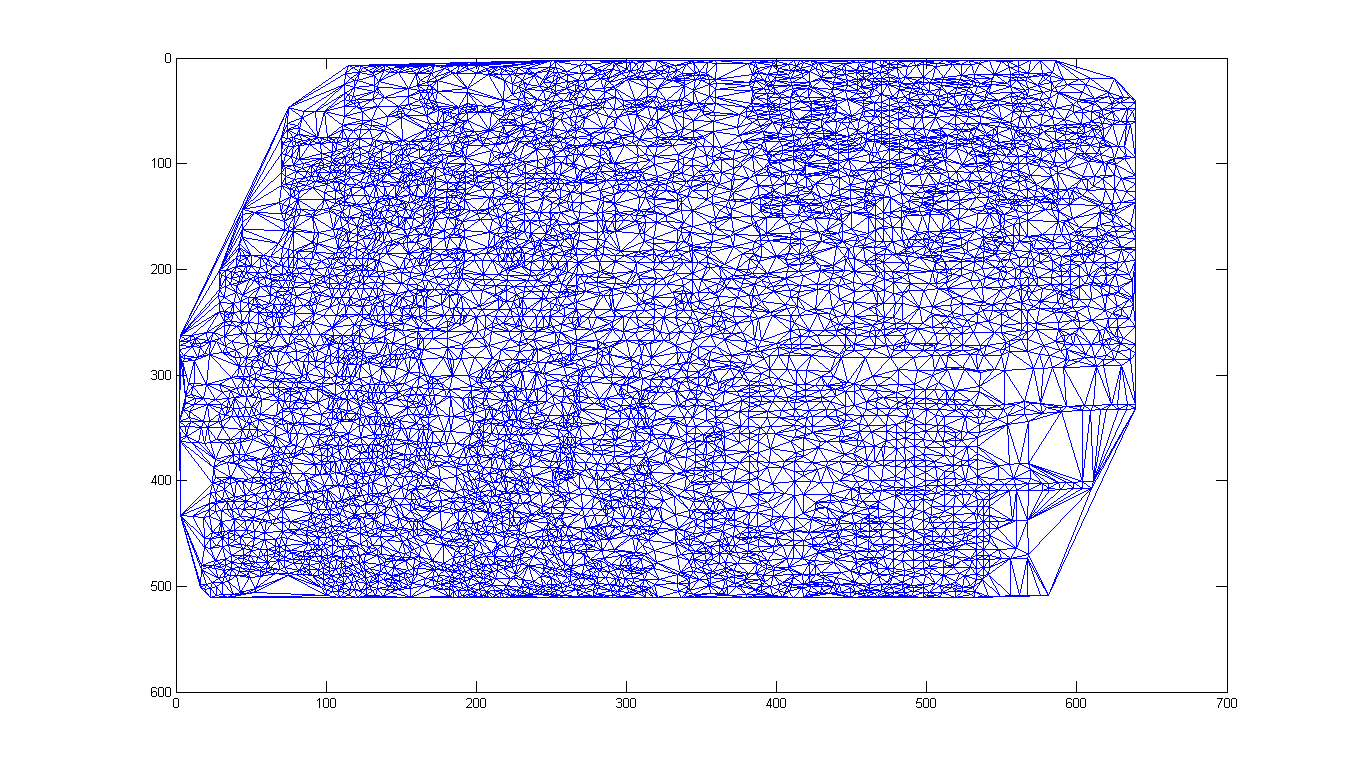}
		\caption{Graph representation}
		\label{fig:max_delauny}
	\end{subfigure}
	\caption{Maximum responses in the set of images. In \ref{fig:max_magni} the maximum magnitude for each location collected in the image is shown (black means no measurement collected) and the corresponding depth values in \ref{fig:max_depth}. \ref{fig:max_locmax} shows the local maxima (pixels increased for visualization) of the maximum magnitude map \ref{fig:max_magni} on which a Delaunay triangulation is applied resulting in a graph representation \ref{fig:max_delauny}.}
	\label{fig:locmax_graph}
\end{figure}
Figure~\ref{fig:Focus_measurement} shows the first step of the algorithm for one input image~\ref{fig:focus_input}. We used the canny edge filter~\cite{canny1986computational} with the first derivative of a Gaussian as kernel ($N'(x,y) = \frac{1}{2\pi\sigma^2}e^{\frac{x^2 + y^2}{2\sigma ^2}} \frac{\partial}{\partial x} \frac{\partial}{\partial y}$). The response (magnitude) of the convolution with this kernel is visualized in the Figure~\ref{fig:focus_mag}. For $\sigma$, which is the standard deviation of the Gaussian, we used $\sqrt{2}$. 
After adaptive threshold selection (95\% are not edges) and nonmaximum suppression of the Canny edge filter, we use the resulting edges (Figure~\ref{fig:focus_edges}) as filter mask. In other words, only magnitude values assigned to a valid edge are allowed to pass. The stored magnitude responses for the input image are shown in Figure~\ref{fig:focus_mag_final}. The idea behind this step is to restrict the amount of information gathered per image, consequently reducing the impact of noise on the algorithm. These two parameters ($\sigma$ and non-edge ratio) are the only variables of the proposed method.

\subsection{Graph Representation}
After in each image the focus measure was applied and filtered, the maximum along z of each location is collected in a maximum map (Figure~\ref{fig:max_magni}, Equation~\ref{eq:max_map}).
\begin{equation}
M(x,y)=max_z(V(x,y,z))
\label{eq:max_map}
\end{equation}
\begin{equation}
D(x,y)=\begin{cases}z&\text{M(x,y) $\in$ V(x,y,z)}\\0&\text{M(x,y)=0}\end{cases}
\label{eq:max_depth_map}
\end{equation}
Equation~\ref{eq:max_map} calculates the maximum map $M$ (Figure~\ref{fig:max_magni}) where $V$ represents the volume of filtered focus measures (one for each image). The coordinates $x,y$ correspond to the image pixel location, and $z$ is the image index in the input image. Equation~\ref{eq:max_depth_map} is the corresponding depth or z-index map $D$ (Figure~\ref{fig:max_depth}) where an image set position is assigned to each maximum value.

In Figure~\ref{fig:max_magni} and its corresponding depth map~\ref{fig:max_depth}, it can be seen that not every pixel has a depth estimation. Additionally, most of the collected edges have traces, meaning that the edge was collected in images recorded with different focal length. The trace occurs because changes in focal length induce a scaling of the field of view of the camera.
\begin{figure}
	\centering
	\begin{subfigure}[b]{0.22\textwidth}
		\includegraphics[width=\textwidth]{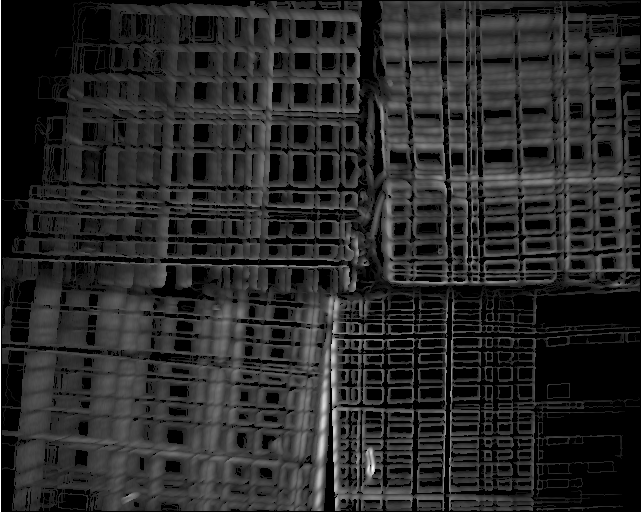}
		\caption{Magnitude trace}
		\label{fig:trace_magni}
		,   \end{subfigure}
	\begin{subfigure}[b]{0.22\textwidth}
		\includegraphics[width=\textwidth]{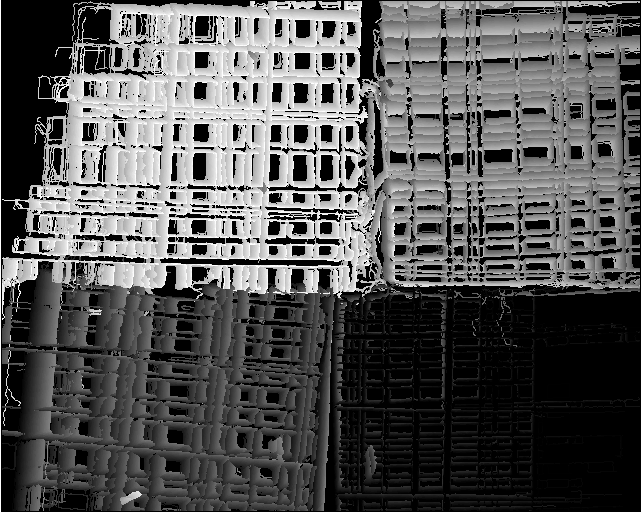}
		\caption{Depth trace}
		\label{fig:trace_depth}
	\end{subfigure}
	\caption{Maximum magnitude responses (\ref{fig:trace_magni}) and the assigned depth index (\ref{fig:trace_depth}) in the set of images. In comparison to figure~\ref{fig:locmax_graph} were only 19 images in the input set where used, here the set consist of 190 images to show the traces more clear.}
	\label{fig:trace_exp}
\end{figure}
In Figure~\ref{fig:trace_exp}, these traces and their occurrences are shown more clear due to the increased amount of the input image set (190 images). For Figure~\ref{fig:locmax_graph}, we used 19 images in the input set. The bottom right part of Figure~\ref{fig:trace_magni} shows that the occurrence of those traces is not present as strongly as in the other parts. This is due to the lens center (in our setup bottom right) from which the field of view scale impact increases linearly in distance.

The next step of the algorithm is the computation of local maxima (Figure~\ref{fig:max_locmax}) and, based on those, setting up a graph by applying the Delaunay triangulation (Figure~\ref{fig:max_delauny}). The idea behind this step is to abstract the depth measurements, making it possible to estimate the depth of plain surfaces (as long as their borders are present) without the need of specifying a window size. Additionally, this graph is used to assign a set of depth estimations to one edge by identifying connected traces. These values are important because the set of input images does not have to contain the optimal focus distance of an edge. Therefore, the set of depth values belonging to one edge are used to interpolate its depth value.

\begin{figure}
	\centering
	\begin{subfigure}[b]{0.22\textwidth}
		\includegraphics[width=\textwidth]{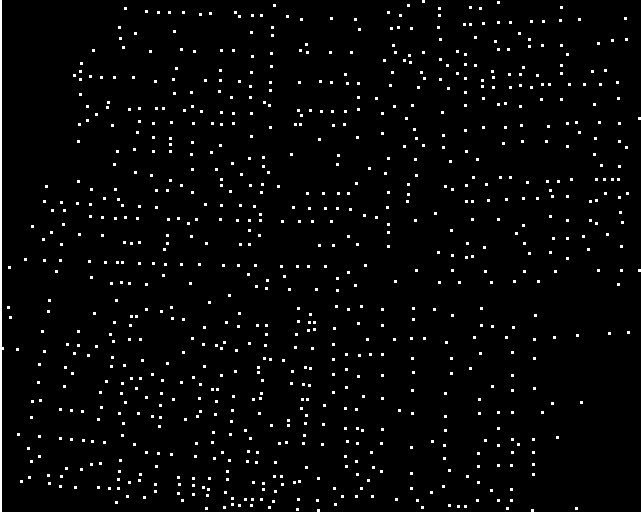}
		\caption{Nodes $G_{max}$}
		\label{fig:g_max}
	\end{subfigure}
	\begin{subfigure}[b]{0.22\textwidth}
		\includegraphics[width=\textwidth]{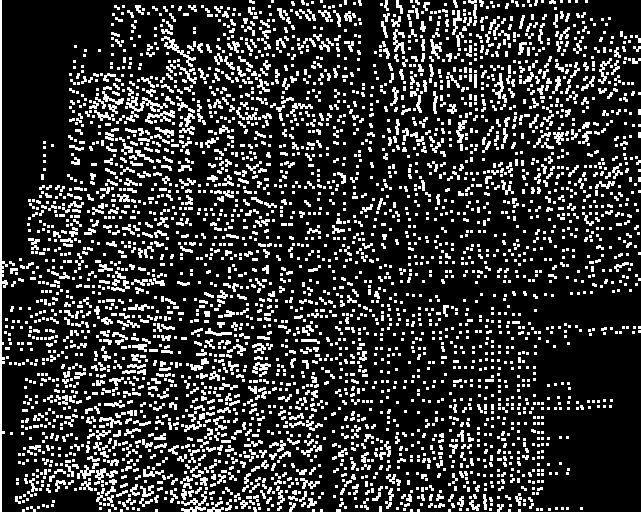}
		\caption{Nodes $G_{nonmax}$}
		\label{fig:g_nonmax}
	\end{subfigure}
	\caption{White dots represent node locations (pixels increased for visualization). In \ref{fig:g_max} the nodes which have an equal or larger magnitude value compared to their connected neighbors in $G_{all}$ are shown. \ref{fig:g_nonmax} show the remaining non maximal nodes of $G_{all}$ after removing those in $G_{max}$. }
	\label{fig:g_max_nonmax}
\end{figure}

The local maxima (\ref{fig:locmax_graph}) are computed based on a eight connected neighborhood on the maximum magnitude map (~\ref{fig:max_magni}). Based on those points, the Delaunay triangulation (\ref{fig:max_delauny}) sets up a graph, where each triple of points creates a triangle if the circumcircle does not contain another point. This graph $G_{all}$ (Figure~\ref{fig:locmax_graph}) contains multiple maxima from the same edge on different depth plains. To separate those, we introduce two types of nodes: a maximal response set $G_{max}$ (Figure~\ref{fig:g_max}) and a non maximal response set $G_{nonmax}$ (Figure~\ref{fig:g_nonmax}).
\begin{equation}
\begin{split}
G_{max}= \forall i \in G_{all}, \forall j \in CN(G_{all},i),\\
V(j) \le V(i)
\end{split}
\label{eq:max_graph}
\end{equation}
\begin{equation}
G_{nonmax}= \forall i \in G_{all}, i \notin G_{max}
\label{eq:nonmax_graph}
\end{equation}
Equation~\ref{eq:max_graph} is used to build the maximal response set $G_{max}$ (Figure~\ref{fig:g_max}) where i is a Node in $G_{all}$ and $CN(G_{all},i)$ delivers all connected neighbors of i. Therefore only nodes with an equal or higher magnitude value compared to their connected neighbors belong to $G_{max}$. $G_{nonmax}$ (Figure~\ref{fig:g_nonmax}) consists of all nodes in $G_{all}$ which are not in $G_{max}$ and specified in equation~\ref{eq:nonmax_graph}.

\subsection{Node Correspondence Collection}
\begin{algorithm}
	\begin{algorithmic}
		\Require{$G_{all}, G_{max}, G_{nonmax}, V$}
		\Statex
		\Function{$Select_{correspondences}$}{$G_{all}, G_{max}, G_{nonmax}$}
		\For{ $a \in G_{max}$}
		\For{ $b \in CN(G_{all},a)$, $b \in G_{nonmax}$}
		\If{$D(a) \neq D(b)$}
		$add(CAN(a),b)$
		\EndIf
		\For{ $c \in CN(G_{all},b)$ AND $c \in G_{nonmax}$}
		\If{$D(a) \neq D(c)$}
		$add(CAN(a),c)$
		\EndIf
		\EndFor
		\EndFor
		\For{ $ z \in V(a)$, $V(a,z)>0$}
		\If{$D(a) \neq z$}
		$add(CAN(a),z)$
		\EndIf
		\EndFor
		\EndFor
		\State \Return{$CAN$}
		\EndFunction
	\end{algorithmic}
	\caption{Algorithm for candidate selection where $CAN(a)$ are all candidates for node a, $CN(a)$ are the connected neighbors to node a, $V$ the set of focus measure responses for each input frame, $z$ the frame index, $D(a)$ the depth index of node a, $G_{all}$ all local maxima, $G_{max}$ all maximal nodes and $G_{nonmax}$ all not maximal nodes.}
	\label{algo:collect_can}
\end{algorithm}
Algorithm~\ref{algo:collect_can} performs the candidate selection. Candidates are possible node correspondences and marked by $CAN(a)$, where a is the index node for the assignment. For each node in $G_{max}$, connected nodes in $G_{nonmax}$ with a different depth value are collected. Since we want to collect all nodes that could possibly build a line over a trace and the maximum could be the last or first measurement, we have to collect the connected nodes to the node from $G_{nonmax}$ too. In case the node from $G_{max}$ is close to the lens center, where the scaling has low to none impact, we have to search in the volume of responses ($V$) as well.

After all candidates are collected, each pair has to be inspected to be a possible line trace or, in other words, a valid pair of corresponding focus measures.
\begin{equation}
\begin{split}
COR(a)&= \forall b,c \in CAN(a),\\
& \begin{cases}b \neq c\\D(a) \neq D(b) \neq D(c)\\\vec{ab} \measuredangle \vec{ac} = \pi\end{cases}
\end{split}
\label{eq:collect_cor}
\end{equation}
Equation~\ref{eq:collect_cor} describes the correspondences collection based on the collected candidates ($CAN(a)$) belonging to node a. The equations after the large bracket are the conditions, where $D(a)$ is the depth index of node a and $\vec{ab} \measuredangle \vec{ac}$ is the angle between the two vectors $\vec{ab}$ and $\vec{ac}$. In our implementation, we used $-0.95$ (-1 corresponds to $\pi$) because an image is a solid grid and floating point inaccuracy.

\subsection{Interpolation}
For estimating the real depth of a node in $G_{max}$, we used the three point Gaussian fit technique proposed by Willert and Gharib~\cite{willert1991digital} and first used for depth estimation by Nayar et al.~\cite{nayar1994shape}. The assumed Gaussian is $M=M_{peak}e^{-0.5 \frac{D-\bar{D}}{\sigma}}$ where $M$ is the focus measure response, $D$ the depth, and $\sigma$ the standard deviation of the Gaussian.
This can be rewritten with the natural logarithm $ln(M)=ln(M_{peak})-0.5 \frac{D-\bar{D}}{\sigma}$. $\bar{D}$ is the depth value where the Gaussian has the highest focus measure (mean) and obtained using equation~\ref{eq:depth_inter}.
\begin{equation}
\begin{split}
M^+(a,b)&=ln(M(a))-ln(M(b))\\
M^-(a,b,c)&=M^+(a,b)+M^+(a,c)\\
D^{2-}(a,b)&=D(a)^2 - D(b)^2\\
\Delta D(a,c)&=2|D(a)-D(c)|\\
\bar{D}(a,b,c)&=\frac{M^+(a,c)*D^{2-}(a,b)}{\Delta D(a,c)*M^-(a,b,c)}
\end{split}
\label{eq:depth_inter}
\end{equation}
In equation~\ref{eq:depth_inter}, $a,b,c$ are node triples obtained from $COR(a)$, where $M$ is the focus measure, and $D$ is the depth value (we used the same letters as in equation~\ref{eq:max_map} and \ref{eq:max_depth_map} for simplification and want to note that it is not valid for nonmembers of $G_{all}$, which are obtained through the response volume (\ref{algo:collect_can}) ). 

Since $COR(a)$ can have more than one pair of possible interpolations, we use the median over all possible interpolation values ($ D(a)=Median(\{\bar{D}(a,b,c)\})$, $\forall b,c \in COR(a)$).

\subsection{Rebuild Graph}
\begin{figure}
	\centering
	\includegraphics[width=0.35\textwidth]{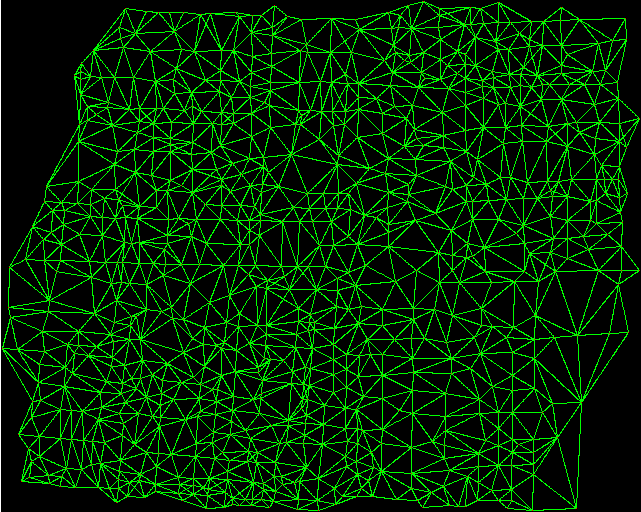}
	\caption{The graph build on $G_{max}$ using Delaunay triangulation.}
	\label{fig:graph_max}
\end{figure}
For using those interpolated nodes in $G_{max}$ as image representation, we have to rebuild the graph. Again we use the Delaunay triangulation with the result shown in Figure~\ref{fig:graph_max}. Due to possible errors from the interpolation or the focus measurement, we apply a median filter on the depth of the node and its neighborhood ($D(a)=Median(\{D(CN(a)), D(a)\})$). This median interpolation is performed interdependently; in other words, the values are stored directly into the depth map $D$, therefore influencing the median filtering of its neighbors. We used this way of median filtering because it delivered slightly better results. A more time-consuming approach would be iteratively determining the median. However, such an iterative approach could lead to oscillation and therefore no convergence.

\subsection{Depth Map Creation}
For depth map creation, the graph in Figure~\ref{fig:graph_max} has to be transformed into a surface. This is done by assigning each pixel in a triangle the weighted value of the depth estimations from the corner nodes. The weights are determined using the distance to each node. The idea behind this is to have linear transitions between regions with different depth values. This makes it more comfortable for the subject to slide over the scene with their gaze, without having an oscillatory effect of the focal length close to region borders.
\begin{figure}
	\centering
	\begin{subfigure}[b]{0.22\textwidth}
		\includegraphics[width=\textwidth]{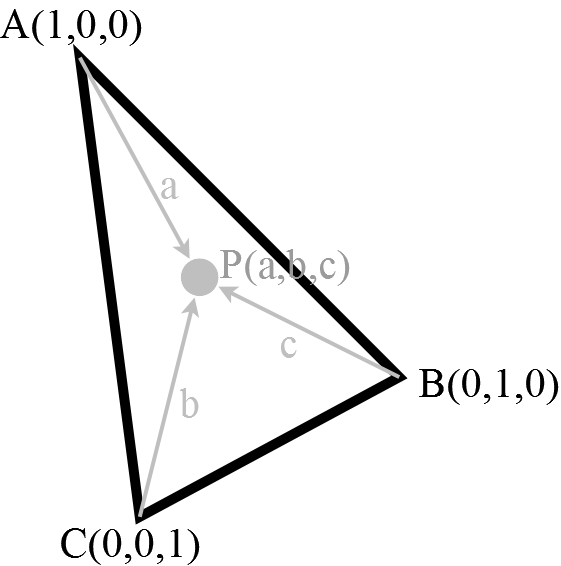}
		\caption{Barycentric}
		\label{fig:barycentric_exp}
	\end{subfigure}
	\begin{subfigure}[b]{0.22\textwidth}
		\includegraphics[width=\textwidth]{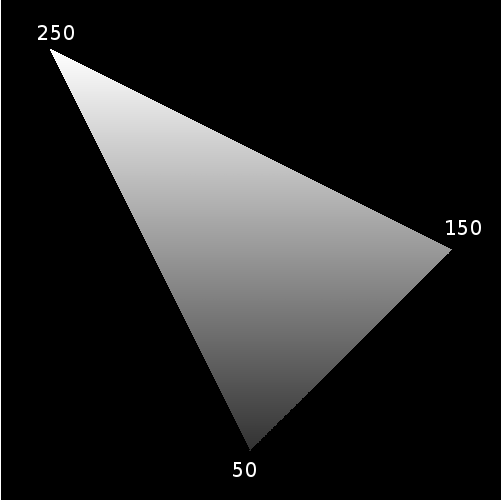}
		\caption{Interpolation}
		\label{fig:barycentric_bsp}
	\end{subfigure}
	\caption{In \ref{fig:barycentric_exp} the barycentric coordinates of a triangle spanned by nodes A,B and C is shown. The gray dot P in the middle of this triangle has coordinates a,b and c which is related to its distance to A, B and C. \ref{fig:barycentric_bsp} shows an exemplary interpolation in such a triangle, where the numbers next to each corner are the intesity value of the corner pixel. }
	\label{fig:barycentric}
\end{figure}
This can be achieved very fast using barycentric coordinates (Figure~\ref{fig:barycentric_exp}) to linearly interpolate (Figure~\ref{fig:barycentric_bsp}) those three values, which is usually applied in computer graphics to 3D models.
\begin{equation}
\begin{split}
P(a,b,c)&\\
a=\frac{\Delta PBC}{\Delta ABC},b=\frac{\Delta PAC}{\Delta ABC}&,c=\frac{\Delta PAB}{\Delta ABC}
\end{split}
\label{eq:barycentric}
\end{equation}
Equation~\ref{eq:barycentric} describes the transformation from Cartesian coordinates to barycentric coordinates, where A,B and C are the corner nodes of a triangle (Figure~\ref{fig:barycentric_exp}), $\Delta$ is the area of the spanned triangle and a,b and c are the barycentric coordinates. An exemplary interpolated triangle can be seen in \ref{fig:barycentric_bsp}.
\begin{equation}
D(P)=a*D(A) + b*D(B) + c*D(C)
\label{eq:barycentric_depth}
\end{equation}
For depth assignment to point $P$ equation~\ref{eq:barycentric_depth} is used where $D$ is the depth value.
\begin{figure}
	\centering
	\begin{subfigure}[b]{0.22\textwidth}
		\includegraphics[width=\textwidth]{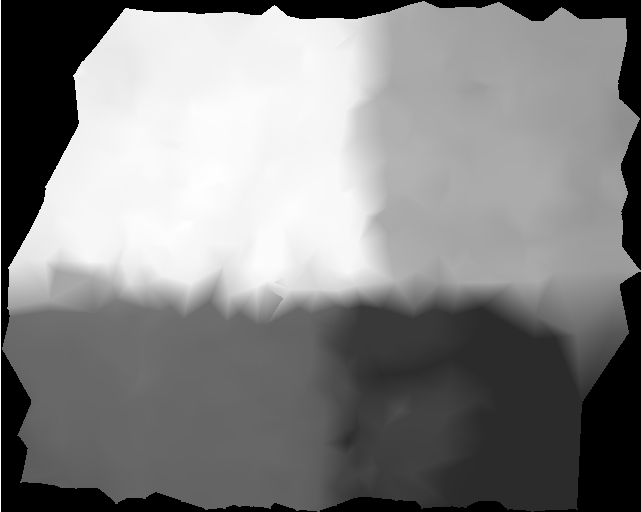}
		\caption{Depth map}
		\label{fig:depth_map}
	\end{subfigure}
	\begin{subfigure}[b]{0.22\textwidth}
		\includegraphics[width=\textwidth]{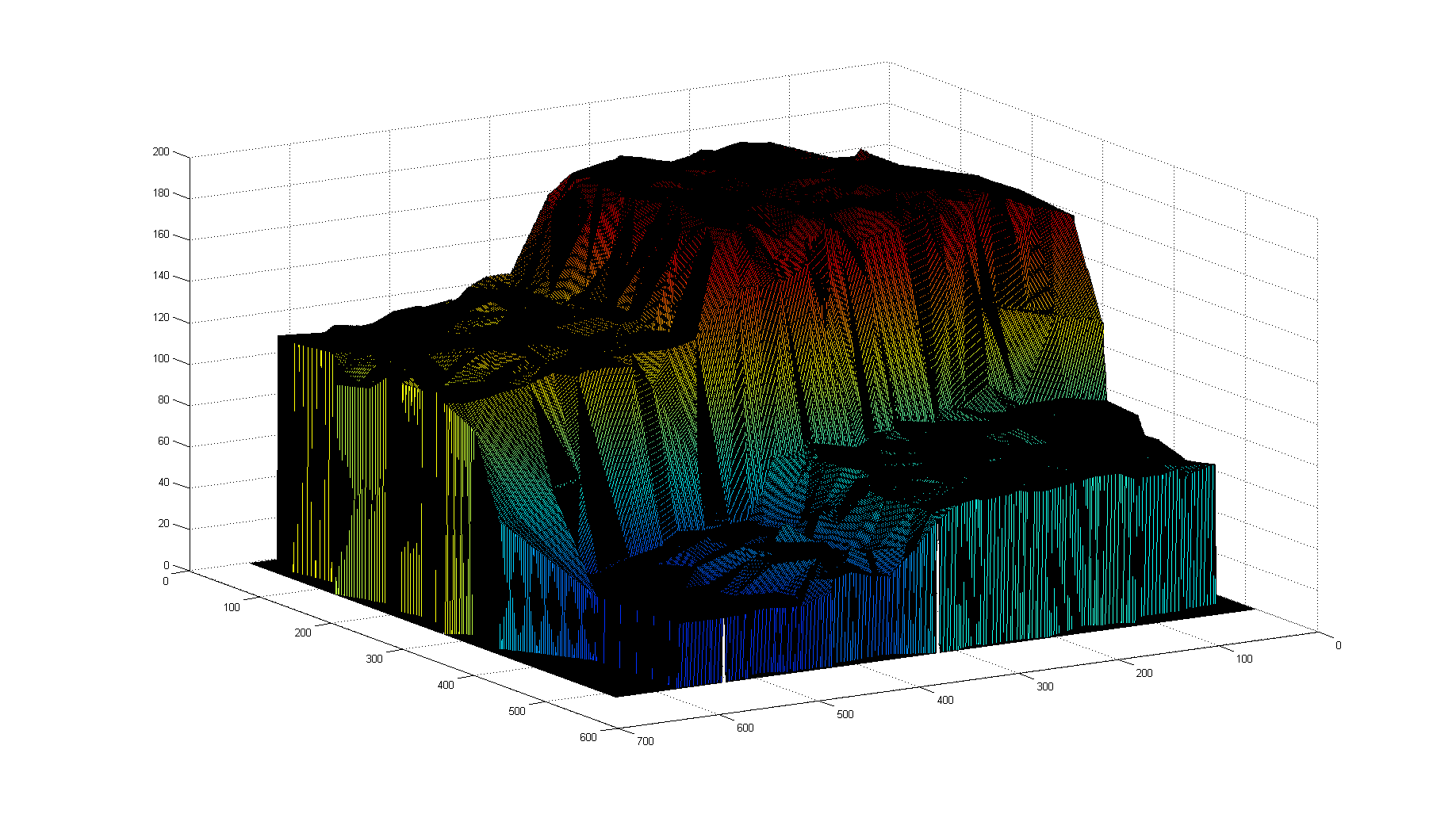}
		\caption{3D model}
		\label{fig:3d_map}
	\end{subfigure}
	\caption{In \ref{fig:depth_map} (normalized) white is closer, dark gray is further away and black means that the depth measure could not estimate a depth value. The 3D model in \ref{fig:3d_map} is generated with the depth map from \ref{fig:depth_map} using matlab. }
	\label{fig:recon_result}
\end{figure}
\begin{figure*}
	\centering
	\includegraphics[width=0.95\textwidth]{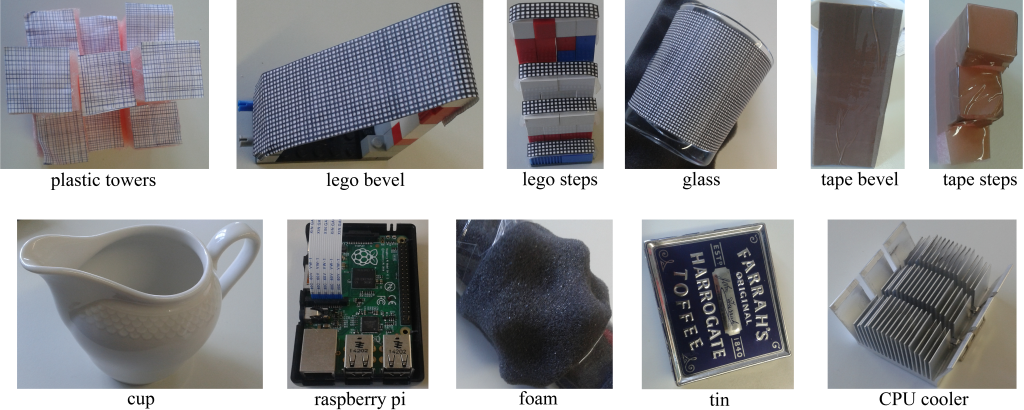}
	\caption{Shows all objects used to generate the datasets. Below each object image stands the title which will be used further in this document.}
	\label{fig:new_datasets}
\end{figure*}
The resulting depth map after linear interpolation of all triangles can be seen in Figure~\ref{fig:depth_map}. In this depth map, white is closer to the camera and dark is further away. Comparing Figure~\ref{fig:depth_map} to Figure~\ref{fig:graph_max} it can be seen that areas for which no enclosing triangle exists are treated as not measurable (black regions in Figure~\ref{fig:depth_map}). If an estimation for those regions is wanted, it is possible to assign those pixels the depth value of the closest pixel with depth information or to interpolate the depth value using barycentric coordinates of the enclosing polygon. The 3D reconstruction based on the depth map from Figure~\ref{fig:depth_map} can be seen in Figure~\ref{fig:3d_map}.

\section{Data sets}
\label{sec:new_data_sets}
In Figure~\ref{fig:new_datasets}, all objects of the new data sets are shown. We scanned each object in 191 steps over the complete range (focal tuning range of 50mm to 120mm~\cite{datasheetel1030}) of the optotune lens. Therefore each object set contains 191 images with a resolution of 640x512.

The objects plastic towers, lego bevel, lego steps, and glass are coated with a grid of black lines which should simplify the depth estimation. For the objects tape bevel and tape steps, we used regular package tape to reduce the focus information for different focal length. Objects cup, raspberry pi, foam, tin, and CPU cooler are real objects where tin and raspberry pi are scanned in an oblique position. All objects except CPU cooler are used for evaluation, whereas the said object is used in limitations because the laminate is interpolated to a flat surface (without modification of the algorithm). This is due to the oppression of the not maximal responses along the laminate (which do not represent real edges).

For evaluation we also used zero motion from Suwajanakorn et al.~\cite{suwajanakorn2015depth} and balcony, alley and shelf from M\"oller et al.~\cite{moeller2015variational}.

\section{Evaluation}

For the evaluation, we used 20 images per object from the set of 191. Those images were equally spread, meaning that the change in focal length between consecutive images is constant. In addition to the objects in Section~\ref{sec:new_data_sets}, we used the data sets provided by Suwajanakorn et al.~\cite{suwajanakorn2015depth} and Moeller et al.~\cite{moeller2015variational}, which are recorded with a professional camera in the real world.

For the evaluation, we did not change any parameter of our algorithm. We used the algorithm variational depth \cite{moeller2015variational} with the parameters as specified by the authors on a GeForce GT 740 GPU. The shape-from-focus (SFF) measures we evaluate are modified gray level variance (GLVM), modified Laplacian~\cite{nayar1989surface} (MLAP), Laplacian in 3D window~\cite{an2008shape} (LAP3d), variance of wavelets coefficients~\cite{yang2003wavelet} (WAVV) and ratio of wavelet coefficients~\cite{xie2006wavelet} (WAVR) as Matlab implementation from Pertuz et al.~\cite{pertuz2013analysis,pertuz2013reliability}, since these are the best performing SFF measures in \cite{pertuz2013analysis}. For optimal parameter estimation, we tried all focus measure filter sizes from 9 to 90 in a stepwise search of 9 as for the median filter size. Additionally, the Gauss interpolation was used. 

\subsection{Algorithm evaluation}
\begin{figure*}
	\centering
	\includegraphics[width=0.95\textwidth]{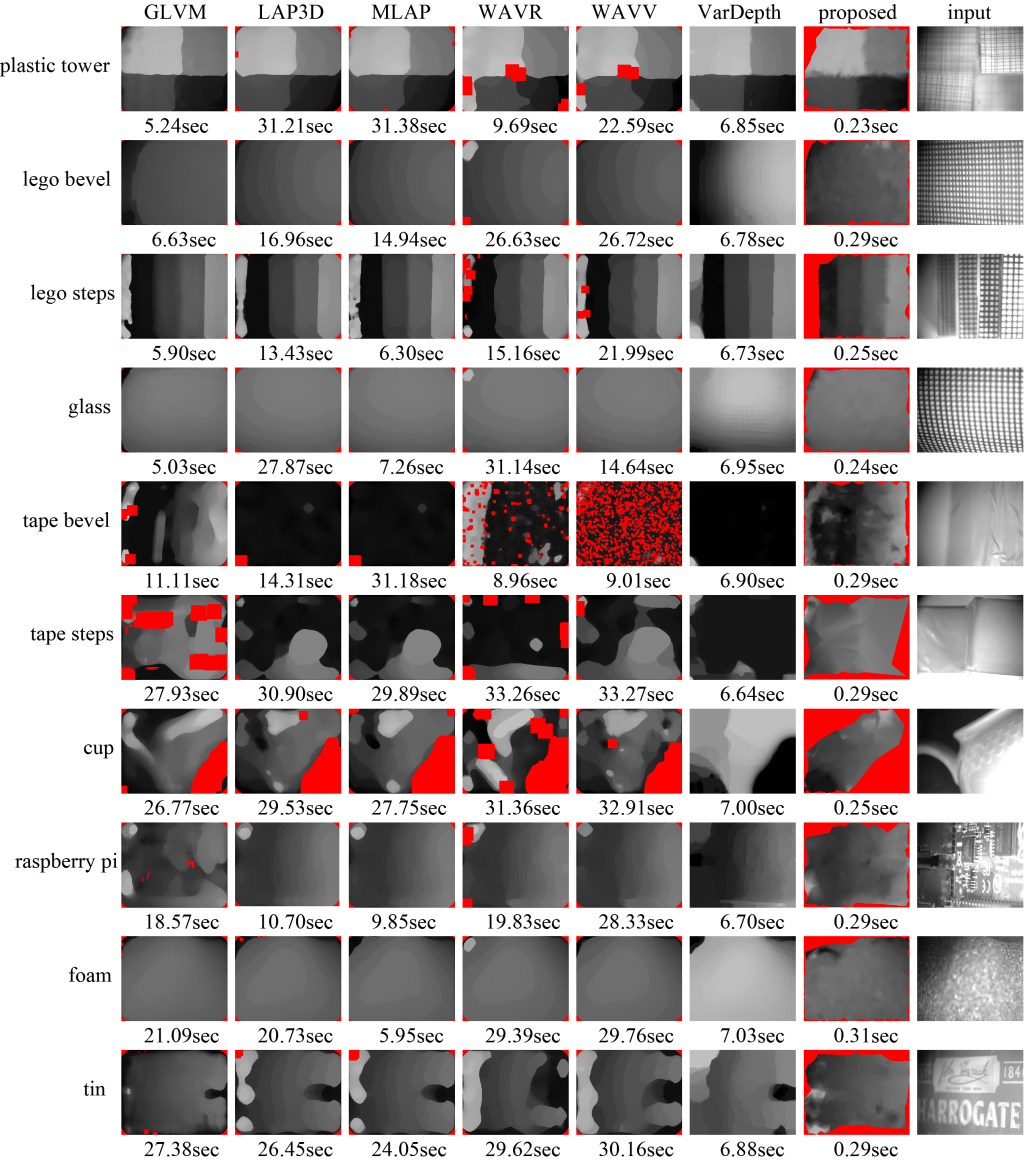}
	\caption{The results on all new data sets in terms of the depth map and the runtime are shown. Red regions represent areas that are marked by the algorithm to be not measurable. Brighter means closer to the camera.}
	\label{fig:eval}
\end{figure*}

\begin{figure*}
	\centering
	\includegraphics[width=0.95\textwidth]{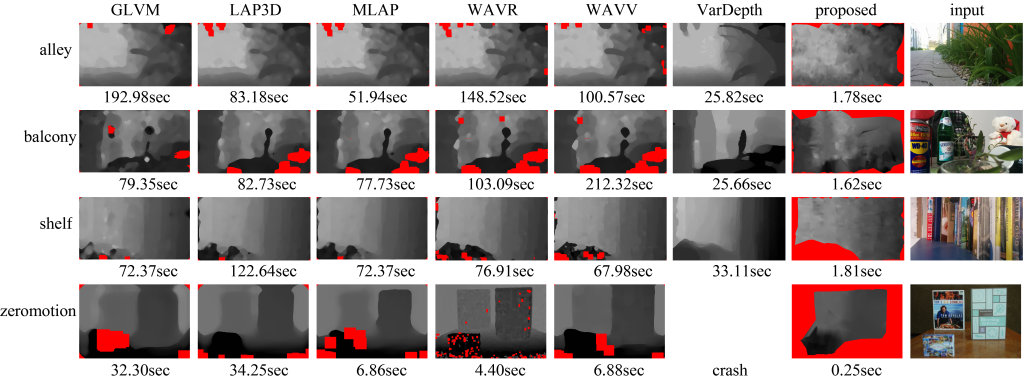}
	\caption{The results on the data sets alley~\protect\cite{moeller2015variational}, balcony~\protect\cite{moeller2015variational}, shelf~\protect\cite{moeller2015variational} and zeromotion~\protect\cite{suwajanakorn2015depth}. The first three data sets have a resolution of 1920x1080, whereas the last one has a resolution of 774x518 pixel. In the top row, each algorithm is named as shown in figure~\ref{fig:eval}. In addition to the depth maps, we show the processing time of each algorithm. Red regions represent areas that were marked by the algorithm as not measurable. Brightness represents the distance to the camera (the brighter, the further away).}
	\label{fig:eval_andere}
\end{figure*}

Since human perception of sharpness varies between subjects, it is very challenging to make exact depth measurements for all objects and to manually label those. Therefore, we decided to show the depth map for each algorithm (proposed, variational depth \cite{moeller2015variational} and SFF) and the timings. In addition, we provide an evaluation against the best index in the image stack as seen by the authors.

Furthermore, in the supplementary material, we provide the parameter settings for SFF. Since SFF was implemented in MATLAB, a comparison with regard to the runtime is not absolutely fair, but we argue, the results are close to what could be reached by a C implementation. For visualization and comparison purposes we normalized the depth maps, based on the input stack with a step size of 10 (between consecutive frames, i.e., the first frame has focus value 1 and the second would have 11). Invalid regions are colored red. The normalization of the algorithm variational depth \cite{moeller2015variational} is always over the complete range because the implementation returns no corresponding scale.

Figure~\ref{fig:eval} shows the results of our proposed method and the state-of-the-art. The first four rows represent the results of objects, where the surface is marked with a grid. The purpose here is to have a comparison based on more objects. For the plastic tower, the irregular transitions between the four regions are due to the triangle interpolation and, therefore, a result as accepted. The corresponding 3D map is shown in Figure~\ref{fig:3d_map}. In the third row (lego steps), it can be seen that our method is able to correctly detect not measurable regions. The closest result is WAVR with 60 times our runtime.

For the tape bevel, we see a superior performance of GLVM, but our method is closest to its result in comparison to the others. For the tape steps object, our method outperforms related approaches in a fraction of runtime. The most difficult object in our data set is the cup, which contains a big reflection and only a small part is textured. The other methods estimate the rim to be closer to the camera than the body of the cup which (is not correct, our method labels large parts as not measurable). The valid region is estimated appropriately with a negligible error (white spot).

The Raspberry PI is estimated correctly by all methods except for GLVM. For foam, all methods perform well. The last object of the new data set is a tin. The two bright spots on the left side of the result of our method are due to dust particles on the lens which get sharp on a closer focal length. The best performing algorithm for the tin is GLVM.

Figure~\ref{fig:eval_andere} presents the results on the data sets provided by Suwajanakorn et al.~\cite{suwajanakorn2015depth} and Moeller et al.~\cite{moeller2015variational}. In comparison to the related algorithms, the results achieved by our approach are not as smooth. For the data set balcony~\cite{moeller2015variational} our algorithm did not get the centered leaf correctly but the depth approximation of the remaining area is comparable to that achieved by the state of the art, while our result does not look as smooth. For the shelf~\cite{moeller2015variational} and zeromotion~\cite{suwajanakorn2015depth} datasets, our algorithm performs better because it detects the not measurable regions. For the algorithm variational depth \cite{moeller2015variational}, it was not possible to provide a depth map for zeromotion from \cite{suwajanakorn2015depth} because of the algorithm crashes (we tried to crop the center out of the image with 640x512 and 640x480 but it is still used to crash).

The purpose of this evaluation is not to show an algorithm which is capable of estimating correct depth maps for all scenarios but to show that our algorithm can produce comparable results in a fraction of runtime without changing any parameter.

We provide all depth maps in the supplementary material and as a download, together with a Matlab script to show them in 3D.

\subsubsection{Best index evaluation}

\setlength{\tabcolsep}{1.5mm}
\begin{table}[h]
	\centering
	\includegraphics[width=0.25\textwidth]{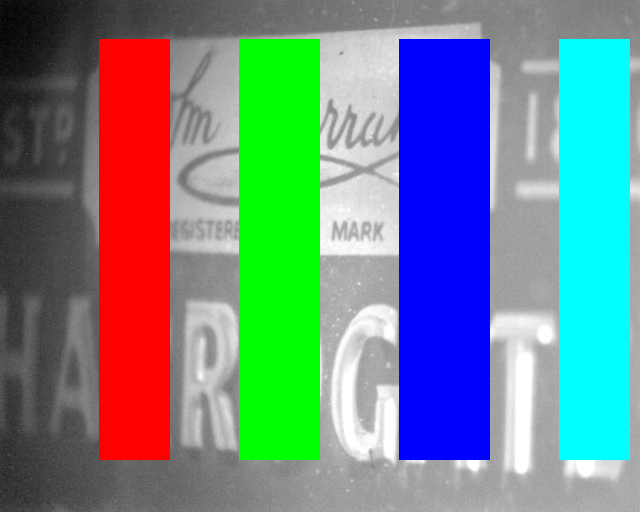}
	\begin{tabular}{c|cccc}
		\textbf{Method} & \textbf{R1} & \textbf{R2} & \textbf{R3} & \textbf{R4}\\ \cline{1-5}
		GLVM & 3.40 &3.88 &6.76 &17.33 \\ \cline{1-5} 
		LAPM & 6.06 &3.80 &11.48 &21.92 \\ \cline{1-5} 
		LAP3 & 12.01 &3.87 &13.47 &22.02 \\ \cline{1-5} 
		WAVV & 19.10 &4.33 &19.87 &30.87 \\ \cline{1-5} 
		WAVR & 27.15 &4.59 &32.53 &57.54 \\ \cline{1-5} 
		VARDEPTH & 28.25 &18.14 &10.29 &35.22 \\ \cline{1-5} 
		Proposed & 12.96 &4.07 &6.96 &5.61 \\ \cline{1-5} 
	\end{tabular}
	\caption{Results for the data set tin. The values represent the mean absolute errors over the marked regions. The regions are named R1-4 and visualized as red for R1, green for R2, blue for R3 and cyan for R4.}
	\label{tbl:eval_blechdose}
\end{table}

\begin{table}[h]
	\centering
	\includegraphics[width=0.25\textwidth]{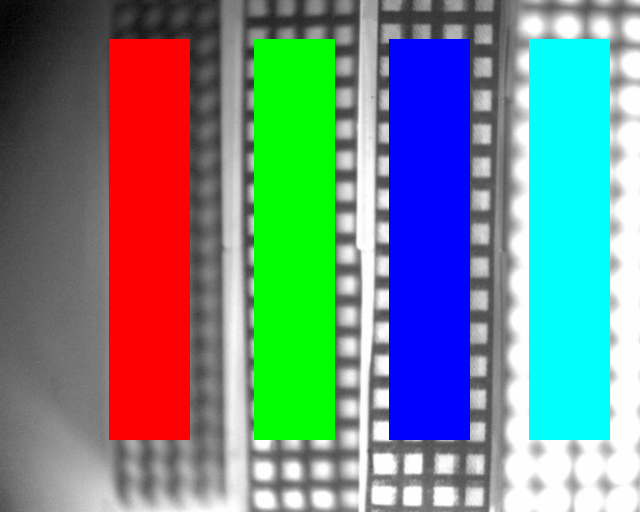}
	\begin{tabular}{c|cccc}
		\textbf{Method} & \textbf{R1} & \textbf{R2} & \textbf{R3} & \textbf{R4}\\ \cline{1-5}
		GLVM & 8.26 &4.65 &3.78 &14.86 \\ \cline{1-5} 
		LAPM & 12.85 &5.28 &3.75 &11.68 \\ \cline{1-5} 
		LAP3 & 12.77 &5.27 &3.61 &11.80 \\ \cline{1-5} 
		WAVV & 12.98 &5.33 &3.57 &11.77 \\ \cline{1-5} 
		WAVR & 14.13 &4.92 &3.09 &11.76 \\ \cline{1-5} 
		VARDEPTH & 19.61 &0.72 &2.68 &12.98 \\ \cline{1-5} 
		Proposed & 3.67 &4.14 &5.28 &12.90 \\ 
	\end{tabular}
	\caption{Results for the data set lego steps. The values represent the mean absolute errors over the marked regions. The regions are named R1-4 and visualized as red for R1, green for R2, blue for R3 and cyan for R4.}
	\label{tbl:eval_lego}
\end{table}

\begin{table}[h]
	\centering
	\includegraphics[width=0.25\textwidth]{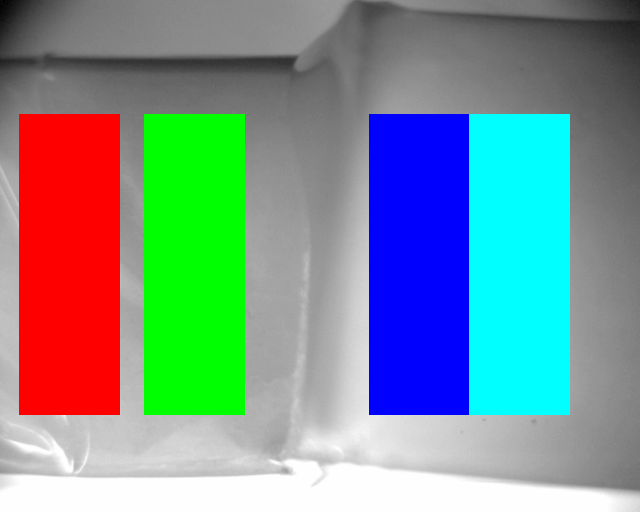}
	\begin{tabular}{c|cccc}
		\textbf{Method} & \textbf{R1} & \textbf{R2} & \textbf{R3} & \textbf{R4}\\ \cline{1-5}
		GLVM & 21.56 &14.05 &25.83 &2.18 \\ \cline{1-5} 
		LAPM & 45.86 &62.75 &39.90 &63.64 \\ \cline{1-5} 
		LAP3 & 45.86 &62.75 &39.90 &63.64 \\ \cline{1-5} 
		WAVV & 34.84 &59.93 &24.33 &43.74 \\ \cline{1-5} 
		WAVR & 56.53 &65.62 &95.51 &112.41 \\ \cline{1-5} 
		VARDEPTH & 52.89 &61.42 &104.43 &109.43 \\ \cline{1-5} 
		Proposed & 12.39 &19.14 &6.27 &15.93 \\ 
	\end{tabular}
	\caption{Results for the data set tape steps. The values represent the mean absolute errors over the marked regions. The regions are named R1-4 and visualized as red for R1, green for R2, blue for R3 and cyan for R4.}
	\label{tbl:eval_plain}
\end{table}

\begin{table}[h]
	\centering
	\includegraphics[width=0.25\textwidth]{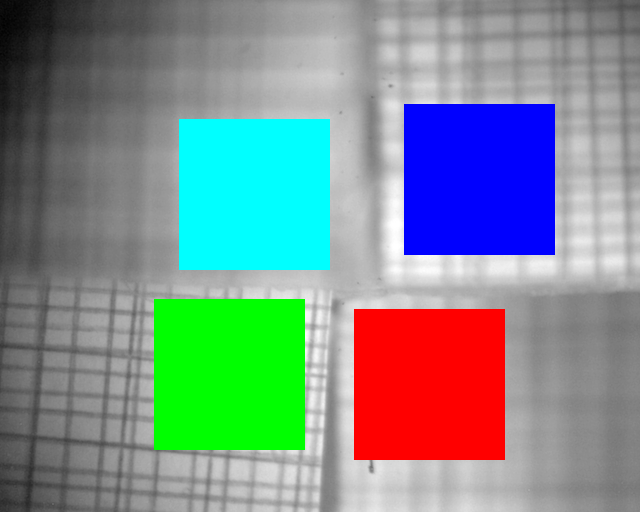}
	\begin{tabular}{c|cccc}
		\textbf{Method} & \textbf{R1} & \textbf{R2} & \textbf{R3} & \textbf{R4}\\ \cline{1-5}
		GLVM & 6.26 &2.08 &2.37 &8.06 \\ \cline{1-5} 
		LAPM & 4.70 &2.29 &3.20 &10.82 \\ \cline{1-5} 
		LAP3 & 4.74 &2.20 &3.22 &10.73 \\ \cline{1-5} 
		WAVV & 9.74 &2.96 &3.75 &13.35 \\ \cline{1-5} 
		WAVR & 7.57 &3.79 &3.95 &13.31 \\ \cline{1-5} 
		VARDEPTH & 8.35 &2.06 &4.24 &12.79 \\ \cline{1-5} 
		Proposed & 3.27 &4.02 &2.57 &2.70 \\ 
	\end{tabular}
	\caption{Results for the data set plastic tower. The values represent the mean absolute errors over the marked regions. The regions are named R1-4 and visualized as red for R1, green for R2, blue for R3 and cyan for R4.}
	\label{tbl:eval_plastik}
\end{table}

For evaluation against the best index in the image stack, we selected a region and an index in which the authors see this region as best focused. It has to be mentioned that we recorded for each data set 191 images with different focal length. For the depth map reconstruction, we used 19 equally spaced images. For evaluation we used the image stacks from "lego steps", "plastic tower", "tape steps" and "tin" (see Figure~\ref{fig:eval}). The same parameters as for the images in Figure~\ref{fig:eval} and~\ref{fig:eval_andere} are used.

The tabels~\ref{tbl:eval_blechdose}, \ref{tbl:eval_lego}, \ref{tbl:eval_plain}, and \ref{tbl:eval_plastik} show the mean absolute error ($\frac{1}{n}|v_i - v_{gt}|$) for the data sets "tin", "lego steps", "tape steps", and "plastic tower", respectively. Over each table we placed an image for the specific data set with the evaluated regions marked by different colors. Regions without depth estimation (marked red in Figure~\ref{fig:eval}) are excluded in the calculation of the mean absolute error. Our methods shows similar performance to the state of the art with less computational time.

\section{Limitations}

\begin{figure}[h]
	\centering
	\begin{subfigure}[b]{0.15\textwidth}
		\includegraphics[width=\textwidth]{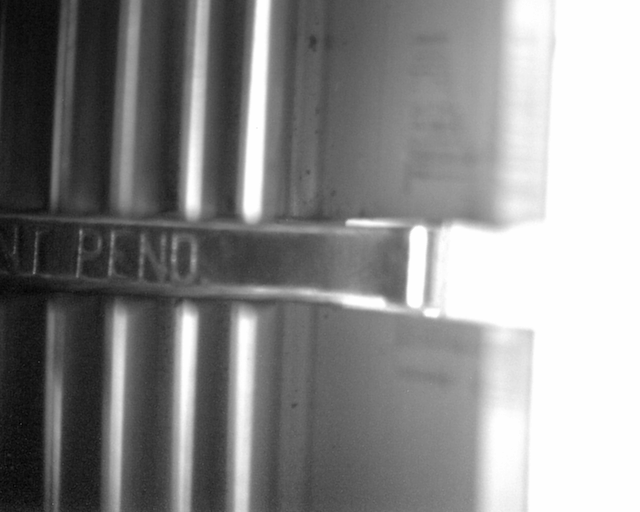}
		\caption{Input}
		\label{fig:input}
	\end{subfigure}
	\begin{subfigure}[b]{0.15\textwidth}
		\includegraphics[width=\textwidth]{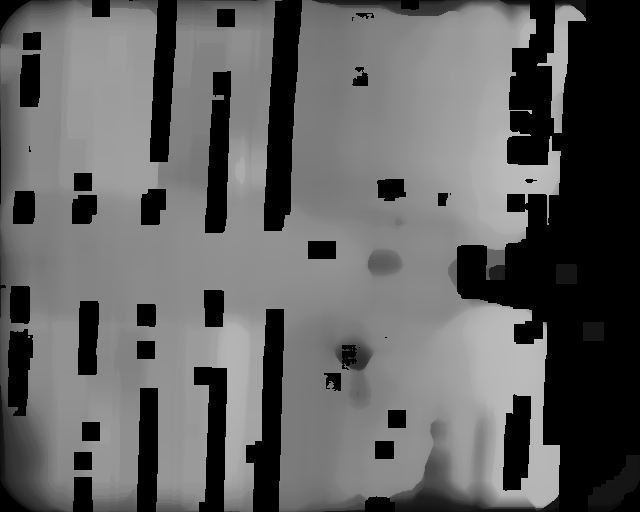}
		\caption{Addapted}
		\label{fig:state_of_the_art}
	\end{subfigure}
	\begin{subfigure}[b]{0.15\textwidth}
		\includegraphics[width=\textwidth]{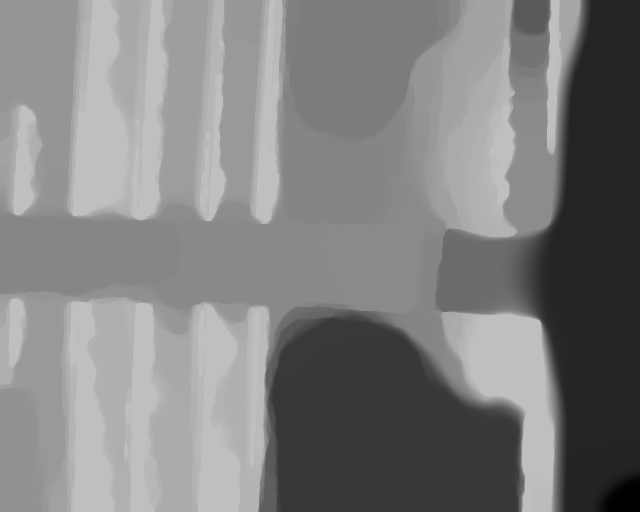}
		\caption{Variational}
		\label{fig:state_of_the_art_var}
	\end{subfigure}
	
	\begin{subfigure}[b]{0.22\textwidth}
		\includegraphics[width=\textwidth]{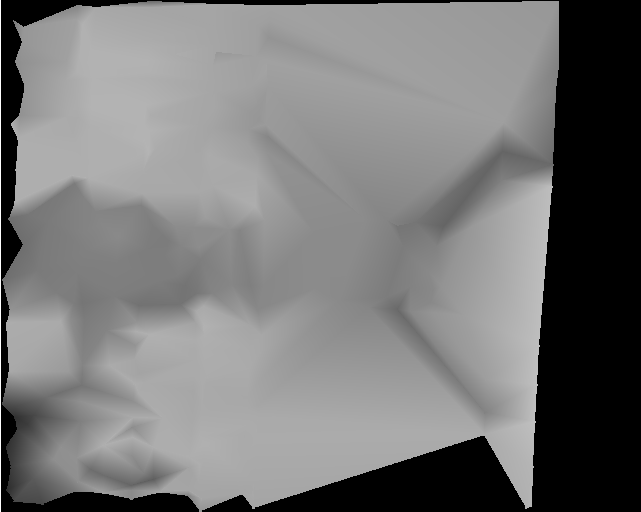}
		\caption{Proposed $G_{max}$}
		\label{fig:std_way}
	\end{subfigure}
	\begin{subfigure}[b]{0.22\textwidth}
		\includegraphics[width=\textwidth]{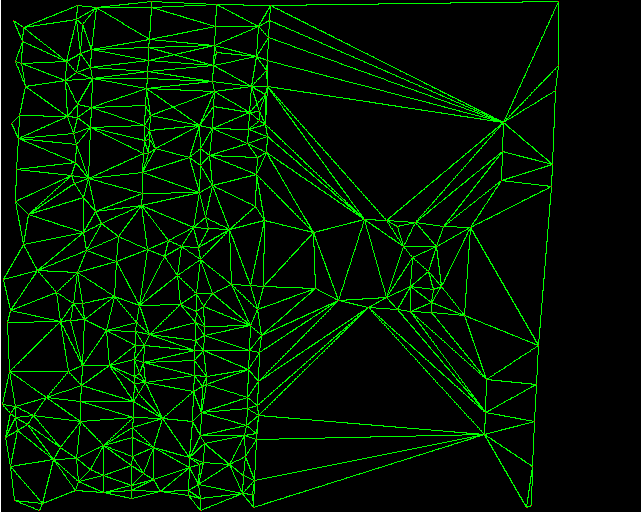}
		\caption{$G_{max}$ graph}
		\label{fig:std_way_graph}
	\end{subfigure}
	
	\begin{subfigure}[b]{0.22\textwidth}
		\includegraphics[width=\textwidth]{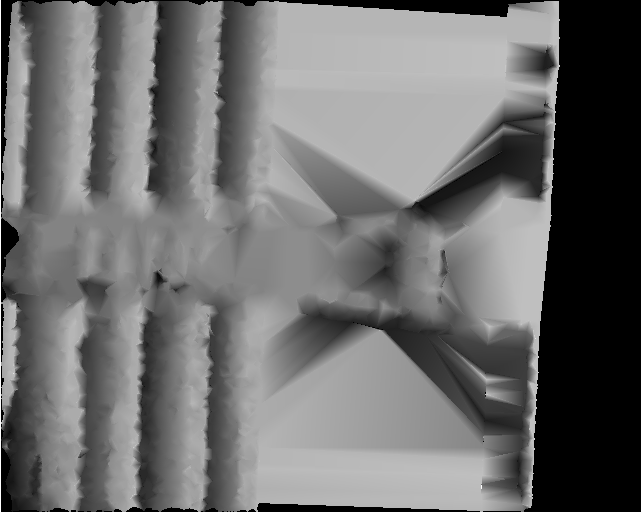}
		\caption{Proposed $G_{all}$}
		\label{fig:nonmax}
	\end{subfigure}
	\begin{subfigure}[b]{0.22\textwidth}
		\includegraphics[width=\textwidth]{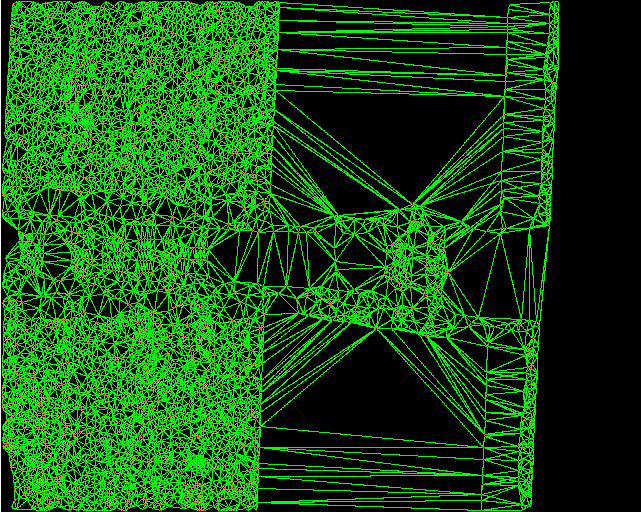}
		\caption{$G_{all}$ graph}
		\label{fig:nonmax_graph}
	\end{subfigure}
	\caption{In \ref{fig:input} one image of the CPU cooler set can be seen. \ref{fig:state_of_the_art} is the result of SFF where we set the optimal parameters (modified gray level variance, filter size 18, median filter size 63, Gauss interpolation, runtime 7.06sec), where black means zero reliability. In \ref{fig:state_of_the_art_var} the result of the variational depth \protect\cite{moeller2015variational} with default parameters is shown (runtime 6.91sec on GPU). \ref{fig:std_way} is the output of the proposed method and the corresponding $G_{max}$ graph in \ref{fig:std_way_graph} (runtime 283ms). Result of the proposed method withy not removing non maximal \ref{fig:nonmax} and the corresponding graph \ref{fig:nonmax_graph} (runtime 296ms).}
	\label{fig:limitations}
\end{figure}

The algorithm is capable of determining plain surfaces if valid measures surrounding this surface are present. This advantage comes at the same time with the drawback that a nonexisting surface is interpolated in case of invalid measures in a region. This effect is depicted in Figure~\ref{fig:std_way}, where the laminate of the CPU cooler is interpreted as a plain surface. The proposed algorithm could determine the lower path in the center of the image but, as can be seen in Figure~\ref{fig:std_way_graph}, all $G_{max}$ nodes are on the top of the laminate resulting in a wrong interpretation.

For a better reconstruction of the CPU cooler, we adapted the proposed algorithm by simply not removing the $G_{nonmax}$ nodes for the second graph building step (meaning we used all nodes in $G_{all}$). As can be seen in Figure~\ref{fig:nonmax}, the resulting depth map got the laminate correctly but still fails for the right-skewed part, which is because there was no valid focus measurement. This can be seen in the graph shown in Figure~\ref{fig:nonmax_graph}.

The disadvantages of using all nodes are the slightly increased runtime (296ms instead of 283ms) due to additional costs introduced by interpolation. Additionally, some nodes are faulty because they either belong to an edge trace or are an invalid measurement. To avoid this, an iterative median filter over should be applied resulting in even higher runtimes. Another approach for improvement would be filtering nodes which have been selected for interpolation. However, this would not reduce the number of faulty measurements.

For comparison, we used the modified gray level variance with filter size 18, median filter size 63 and Gaussian interpolation, which delivered the best result. The runtime was 7.06sec (Matlab implementation from Pertuz et al.~\cite{pertuz2013analysis,pertuz2013reliability}) and is shown in Figure~\ref{fig:state_of_the_art}. The black regions are added to the depth map because of zero reliability, otherwise, the focus measure has everywhere a result. The algorithm variational depth \cite{moeller2015variational} produced the result in Figure~\ref{fig:state_of_the_art_var} with default parameters and a runtime 6.91sec on a GeForce GT 740 GPU. 

For calculation, we used 20 images as for the evaluation and did not change the Canny parameters for the proposed algorithm.

\section{Conclusion}
In this paper, we showed the use of eye tracking for automatically adapting, focus to the presented image. Due to the real-time capability, our methods can be beneficial in various applications where autofocus facilitates interaction (e.g. surgery, security, human-robot collaboration, etc.). The algorithm proposed here shows similar performance as the state of the art but requires minimal computational resources and requires no parameter adjustment.

In our future work, we will further optimize our method and provide a GPU implementation to make it computationally possible to use more than only one focus measuring method (e.g. GLVM). For evaluation purpose, we also have to increase the data sets by different materials and structures.

	\bibliographystyle{elsarticle-harv}
	\bibliography{references}







\end{document}